\documentclass[11pt,a4paper]{article}
\usepackage[hyperref]{acl2021}
\usepackage{times}
\usepackage{listings}
\usepackage{latexsym}
\usepackage{graphicx}
\graphicspath{ {./images/} }

\usepackage{microtype}

\aclfinalcopy 

\title{Underwater Robotics Semantic Parser Assistant}

\author{Jake Imyak \\
  \texttt{imyak.1@osu.edu} \\\And
  Parth Parekh \\
  \texttt{parekh.86@osu.edu} \\\And
  Cedric McGuire \\
  \texttt{mcguire.389@osu.edu} \\}

\date{December 10th, 2021}

\begin{document}
\maketitle
\begin{abstract}
Semantic parsing is a means of taking natural language and putting it in a form that a computer can understand. There has been a multitude of approaches that take natural language utterances and form them into lambda calculus expressions - mathematical functions to describe logic. Here, we experiment with a sequence to sequence model to take natural language utterances, convert those to lambda calculus expressions, when can then be parsed, and place them in an XML format that can be used by a finite state machine. Experimental results show that we can have a high accuracy model such that we can bridge the gap between technical and nontechnical individuals in the robotics field. 
\end{abstract}

\section{Credits}

Jake Imyak was responsible for the creation of the 1250 dataset terms and finding the RNN encoder/decoder model. This took 48 Hours. Cedric McGuire was responsible for the handling of the output logical form via the implementation of the Tokenizer and Parser. This took 44 Hours. Parth Parekh assembled the Python structure for behavior tree as well as created the actions on the robot. This took 40 Hours. All group members were responsible for the research, weekly meetings, presentation preparation, and the paper. In the paper, each group member was responsible for explaining their respective responsibilities with a collaborative effort on the abstract, credits, introduction, discussion, and references. A huge thanks to our Professor Dr. Huan Sun for being such a great guide through the world of Natural Language Processing.

\section{Introduction}

Robotics is a hard field to master. Its one of the few fields which is truly interdisciplinary. This leads to engineers with many different backgrounds working on one product. There are domains within this product that engineers within one subfield may not be able to work with. This leads to some engineers not being able to interact with the product properly without supervision. 

As already mentioned, we aim to create an interface for those engineers on the Underwater Robotics Team (UWRT). Some members on UWRT specialize in other fields that are not software engineering. They are not able to create logic for the robot on their own. This leads to members of the team that are required to be around when pool testing the robot. This project wants to reduce or remove that component of creating logic for the robot. This project can also be applied to other robots very easily as all of the main concepts are generalized and only require the robots to implement the actions that are used to train the project.

\section{Robotics Background}
\subsection{Usage of Natural Language in Robotics}
Robots are difficult to produce logic for. One big problem that most robotics teams have is having non-technical members produce logical forms for the robot to understand. Those who do not code are not able to manually create logic quickly. 
\subsection{Finite State Machines}
One logical form that is common in the robotics space is a Finite State Machine (FSM). FSMs are popular because they allow a representation to be completely general while encoding the logic directly into the logical form. This means things such as control flow, fallback states, and sequences to be directly encoded into the logical form itself. 

\begin{figure}
        \center{\includegraphics[scale=.3]
        {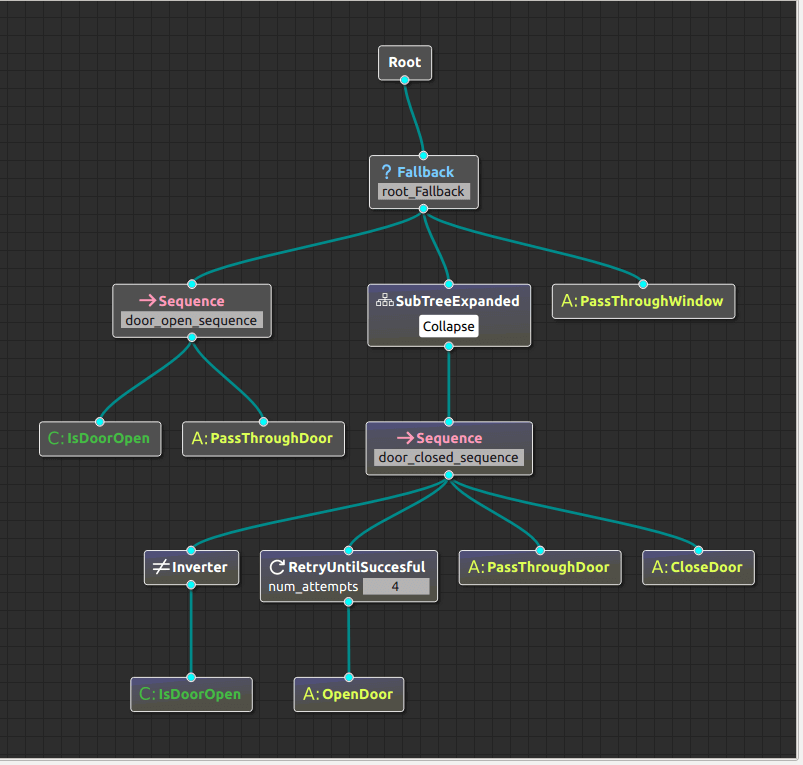}}
        \caption{\label{fig:fsm} A FSM represented in Behaviortree.CPP  \cite{Groot} \cite{BTree} }
\end{figure}

As illustrated in Figure \ref{fig:fsm}, we can easily encode logic into this representation. Since it easily generified, FSM's can be used across any robot which implements the commands that are contained within it. 

\subsection{Underwater Robotics Team Robot}
Since 2016, The Underwater Robotics Team (UWRT) at The Ohio State University has iterated on the foundations of a single Autonomous Underwater Vehicle (AUV) each year to compete at the RoboSub competition. Breaking from tradition, the team decided to take the 2019-2021 school years to design and build a new vehicle to compete in the 2021 competition. Featuring an entirely new hull design, refactored software, and an improved electrical system, UWRT has created its brand-new vehicle, Tempest. \cite{UWRTAbs}
\subsubsection{Vehicle}
Tempest is a 6 Degree of Freedom (DOF) AUV with vectored thrusters for linear axis motion and direct drive heave thrusters. This allows the robot to achieve any orientation in all 6 Degrees of freedom [X, Y , Z, Roll, Pitch, Yaw]. 

\begin{figure}[!htb]
        \center{\includegraphics[scale=.12]
        {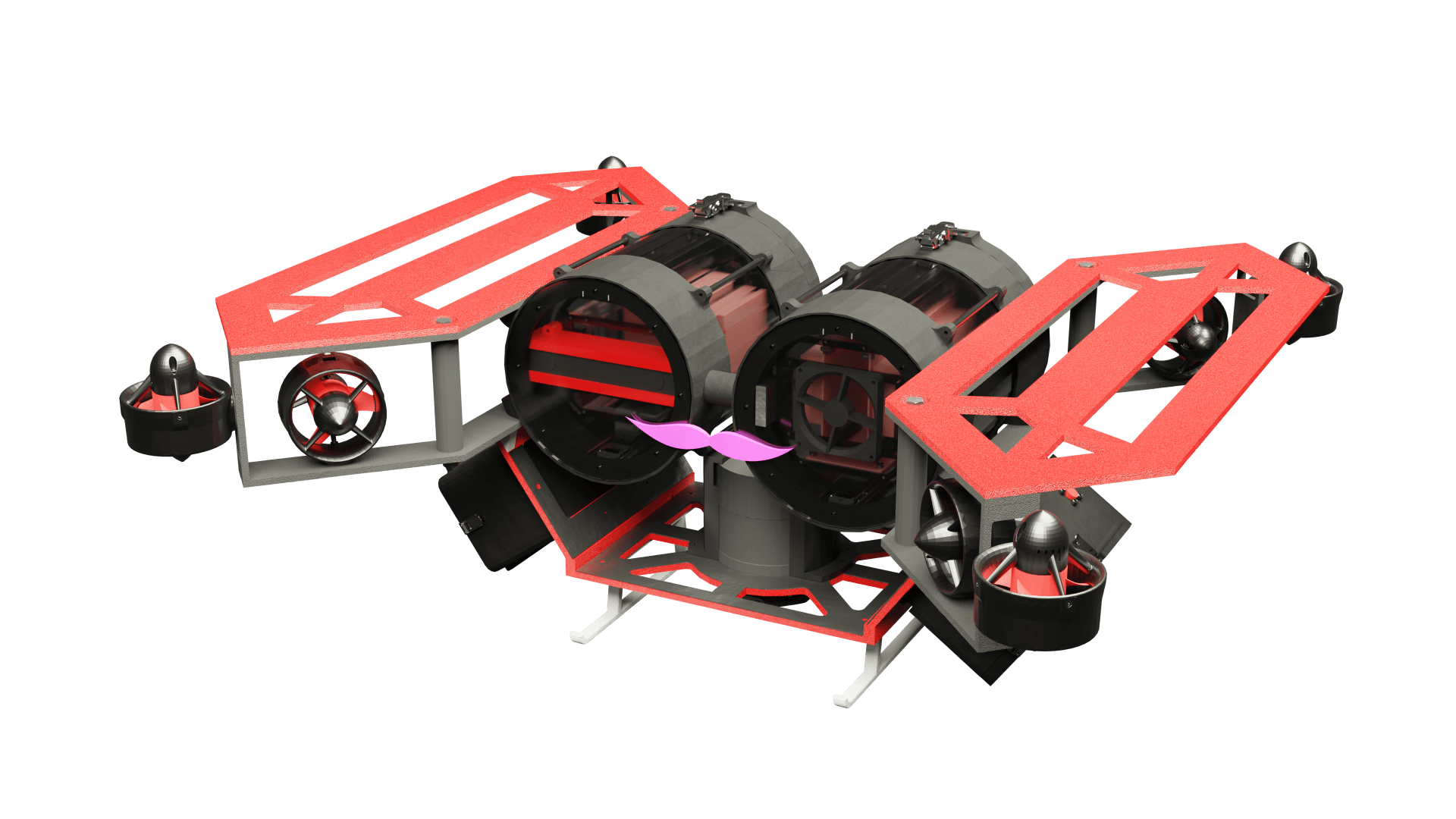}}
        \caption{\label{fig:my-label} A render of Tempest}
\end{figure}

\subsubsection{Vehicle Experience}
With this vehicle, the team has focused on creating a fully fleshed out experience. This includes commanding and controlling the vehicle. One big focus of the team was to make sure that any member, technical or non-technical was able to manage and operate the robot successfully. 

\subsubsection{Task Code System}
A step to fulfill this focus was to change the vehicle's task code system to use the FSM representation. This is done through the library \texttt{\small BehaviorTree.CPP} \cite{BTree}. This generic FSM representation allows for Tempest to use generified logical forms that can be applied to ANY robotic plant as long as that plant implements those commands. This library also creates and maintains a Graphical User Interface (GUI) which allows for visual tracking and creation of FSM trees. Any tree created by the GUI is stored within an XML file to preserve the tree structure.  The structure of the output of the XML syntax is explained within the parser section.

\section{Data}\label{grammar}

A dataset was to be created in order to use natural language utterances to lambda calculus expressions that a parser would be able to recognize to convert to a finite state machine. For reference, the following datasets were considered: the Geoquery set\cite{zettlemoyer2012learning} and General Purpose Service Robotics commands set \cite{GPSR}. The Geoquery dataset provided a foundation for a grammar to follow for the lambda calculus expression such that consistency would hold for our parser. Moreover, the gpsr dataset provided an ample amount of examples and different general purpose robotics commands that could be extended within the dataset we curated. 

The dataset followed the following form: natural language utterance followed by a tab then a lambda calculus expression. The lambda calculus expression is of the form \texttt{\small  ( seq ( action0 ( \$0 ( parameter ) ) )} ... \texttt{\small ( actionN ( \$N ( parameter ) ) )}. The power of the following expression is that it can be extended to N number of actions in a given sequence, meaning that a user can hypothetically type in a very complex string of action and an expression will be constructed for said sequence. Moreover, the format of our dataset allows for it to be extended for any type of robotics command that a user may have. They just need to include examples in the train set with said action and the model will consider it.

\noindent The formal grammar is:

\noindent $<seq>$ : \texttt{\small ( seq ( action ) [ (action) ] )}

\noindent $<action>$ : \texttt{\small actionName [ (parameter ] )}

\noindent $<parameter>$ : \texttt{\small paramName $\lambda$ ( \$n ( n ) )}

The dataset we created had 1000 entries in the training dataset and 250 entries in the test dataset. The size of the vocabulary $|V|=171$ for the input text and $|V|=46$ for the output text, which is similar in vocabulary size to the GeoQuery dataset. The expressions currently increase in complexity in terms of the number of actions within the sequence. A way to extend the complexity of the expressions would make the $<seq>$ tag a nonterminal to chain together nested sequences. The actions within our dataset currently are as follows: \texttt{\small move} (params: x, y, z, roll, pitch, raw), \texttt{\small flatten} (params: num), \texttt{\small say} (params: words), \texttt{\small clean} (params: obj), \texttt{\small bring} (params: val), \texttt{\small find} (params: val), \texttt{\small goal}, and \texttt{\small gate}. The most complex sequence is a string of seven subsequent actions. 

\section{Model}
\label{sec:length}

\subsection{Seq2Seq Model}
We decided to use the model presented in "Language to Logical Form with Neural Attention" \cite{dong2016language}. There was an implementation on GitHub \cite{Avikdelta2018} utilizing Google's Tensorflow library to handle all implementation details of the following model. The part of the paper that was presented was the Sequence to Sequence model with an attention mechanism.  

\begin{figure}[!htb]
        \center{\includegraphics[scale=.5]
        {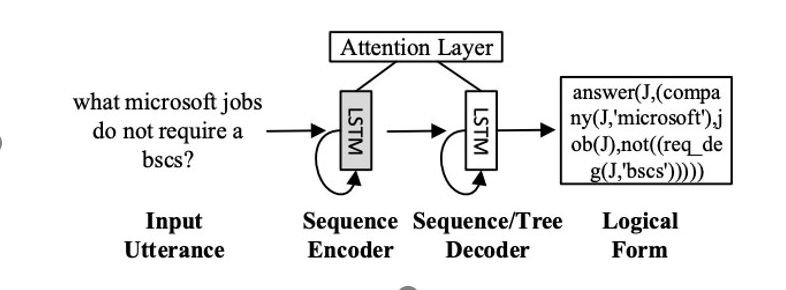}}
        \caption{\label{fig:s2smo} Process of how input natural language are encoded and decoded via recurrent neural networks and an attention mechanism to find the utterance's respective natural language form. (Dong and Lapata, 2016)}
\end{figure}

The model interprets both the input and output from the network as sequences of information. This process is represented in Figure \ref{fig:s2smo}: input is passed to the encoder, then passed through the decoder, and through using the attention mechanism, we can get an output that is a lambda calculus expression. Both of these sequences can be represented as L-layer recurrent neural networks with long short-term memory (LSTM) that are used to take the tokens from the sentences and the expressions we have. The model creates 200 (can be changed to increase and decrease the size of the network) units of both LSTM cells and GRU cells. The GRU cells are used to help compensate for the vanishing gradient problem. These LSTM and GRU cells are used in the input sequence to encode $x_1, ..., x_{q}$ into vectors. Then these vectors are what form the hidden state of the beginning of the sequence in the decoder. Then in the decoder, the topmost LSTM cell predicts the t-th output token by taking the softmax of the parameter matrix and the vector from the LSTM cell multiplied by a one-hot vector used to compute the probability of the output from the probability distribution. The softmax used here is sampled softmax, which only takes into account a subset of our vocabulary V rather than everything to help alleviate the difficulty of finding the softmax of a large vocabulary. 

\subsection{Attention Mechanism}
The model also implemented an attention mechanism to help with the predicted values. The motivation behind the attention mechanism is to use the input sequence in the decoding process since it is relevant information for the prediction of the output token. To achieve this, a context vector is created which is the weighted sums of the hidden vectors in the encoder. Then this context vector is used as context to find the probability of generating a given output. 

\subsection{Training}
To train the model, the objective is the maximize the likelihood of predicting the correct logical form given some natural language expression. Hence, the goal is to minimize the sum of the log probability of predicting logical form a given natural language utterance $q$ summed over all training pairs. The model used the $RMSProp$ algorithm which is an extension of the Adagrad optimizer but utilizes learning rate adaptation. Dropout is also used for regularization which helps out with a smaller datasets to prevent overfitting. We performed 90 epochs.

\subsection{Inference}
To perform inference, the argmax is found of the probability of candidate output given the natural language utterance. Since it is not possible to find the probability of all possible outputs, the probability is put in a form such that a beam search can be employed to generate each individual token of lambda calculus expression to get the appropriate output. 

\section{Results}
With the default parameters set, the Sequence to Sequence model achieved 86.7\% accuracy for exact matches on the test dataset. This is consistent with the model's performance on the Geoquery dataset, achieving 83.9\% accuracy. The test dataset provided contained a 250 entries of similar utterances to the train dataset of various complexities ranging anywhere from one to six actions being performed. There are other methods of evaluating we would like to look into in the future such as computing something such as an F1 score rather than solely relying on exact logical form matching. 

This accuracy for exact logical forms is really important when using the parser. It allows for FSM representation to be easily and quickly built. We were able to build the XML representation and run basic commands on the robot with the model maintaining the order we said them in. 

\section{Logical Form Parser}
The logical form output of our model is sent to a custom parser. The goal of this parser is to translate the output form into BehaviorTree XML files, in which the robot is able to read in as a finite state machine.

\subsection{Tokenizer}
The Tokenizer comprises the initial framework of the parser. It accepts the raw logical form as a String object and outputs a set of tokens in a Python List. These tokens are obtained by looking for separator characters (in our case, a space) present in the logical form and splitting them into an array-like structure. The Tokenizer method permits custom action, parameter, and variable names from the logical form input, thus allowing ease of scalability in implementing new robot actions. Our model's output nature is not able to generate syntactically incorrect logical forms, thus our implementation does not check for invalid tokens and will assume all input is correct. The Tokenizer is stored in a static $Singleton$ class such that it can be accessed anywhere in the program once initialized. It keeps track of the current token (using \texttt{\small getToken()}) and has an implementation to move forward to the next token \texttt{\small skipToken()}. This functionality is important for the object-oriented approach of the parser, discussed in the next section.   

\subsection{Parsing Lambda Calculus Expressions}
The output tokens from the Tokenizer must be interpreted into a proper Python from before they are staged to be turned into XML-formatted robot-ready trees. This is the function of the middle step of the parser, in which a tree of Python objects are built. The parser utilizes an object-oriented approach. As such, we include three objects: $Sequence$, $Action$, and $Parameter$, with each corresponding to an individual member of our custom grammar. The objects orient themselves into a short 3-deep tree, consisting of a $Sequence$ root, $Action$ children, and $Parameter$ grand-children. Each object has its own \texttt{\small parse()} method that will advance the tokenizer, validate the input structure, and assemble themselves into a Python structure to be staged into an XML file. The validations are enforced through our grammar definitions in Section \ref{grammar}. 

\subsubsection{Sequence Object}
The $Sequence$ object is the first object initialized by the parser, along with the root of our action tree. Each $Sequence$ is composed of a list of 0 or more child actions to be executed in the order they appear. The \texttt{\small parseSequence()} method will parse each individual action using \texttt{\small parseSAction()}, all the while assembling a list of child actions for this $Sequence$ object. As of now, $Sequence$ objects are unable to be their own children (i.e. nesting $Sequence$s is not permitted). However, if required, the $Sequence$ object's \texttt{\small parseSequence()} method can be modified to recognize a nested action sequence and recursively parse it.

\subsubsection{Action Object}
$Action$ objects define the title of the action being performed. Similar to $Sequence$, $Action$ objects have an internally stored list, however with $Parameter$ objects as children. There may be any number of parameters, including none. When \texttt{\small parseAction()} method is called, the program validates the tokens and will call \texttt{\small parseParameter()} on each $Parameter$ child identified by the action. 

\subsubsection{Parameter Object}
The $Parameter$ object is a simple object that stores a parameter's name and value. The parser does not have a check for what the name of the parameter is, nor does it have any restrictions to what the value can be. \texttt{\small parseParameter()} searches through the tokens for these two items and stores them as attributes to the $Parameter$ object. This implementation of parameter is scalable with robot parameters and allows any new configuration of parameter to pass by without any changes in the parser as a whole. If a new parameter is needed for the robot, it only has to be trained into the Seq2Seq model on the frontend and into the robot itself on the backend; the Parameter object should take care of it all the same.
\subsection{BehaviorTree Output}
\begin{figure}[!htb]
        \center{\includegraphics[scale=.45]
        {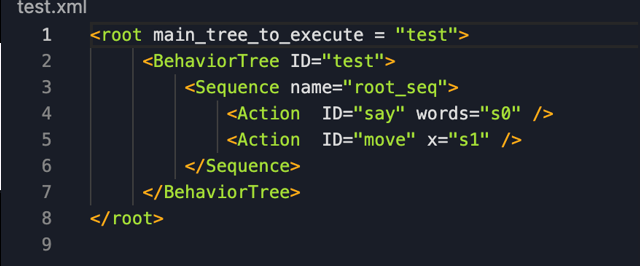}}
        \caption{\label{fig:behavtree} A FSM that was generated from test input through our RNN}
\end{figure}

In the end, the parser outputs an XML file which can be read in to BehaviorTree.CPP \cite{BTree}. An example of this file structure is shown in Figure \ref{fig:behavtree}. 

This file structure is useful because it encodes sequence of actions within it. The leaves of the sequence are always in order. The tree can also encode subtrees into the sequence which we have not implemented yet.

\section{Discussion}
\subsection{Summary}
We learned that semantic parsing is excellent tool at bridging the gap between both technical and nontechnical individuals. The power within semantic parsing with robotics is that any human can automate any task just through using their words. Our dataset is written in a way that just extending the entries with another robot's tasks that use a behavior tree to perform action, that robot's actions can be automated as well. 

\subsection{Future Plans}
Future plans with this project would be to expand the logical flow that can be implemented with BehaviorTree.CPP. As an FSM library, BehaviorTree.CPP implements many more helper functions to create more complicated FSMs. These include things like if statements fallback nodes, and subtrees. This would be a valid expansion of our RNN's logical output and with more time, we could support the full range of features from BehaviorTree.CPP

We would also like to implement a front end user interface to make this service more accessible to anyone who was not technical. Right now, the only means of running our program is through the command line which is not suitable for individuals who are nontechnical. Moreover, including a speak-to-text component to this project would elevate it since an individual would be able to directly tell a robot what commands to do, similar to a human.

\subsection{Source Code}

You can view the source code here: \url{https://github.com/jrimyak/parse\_seq2seq}

\end{document}